\documentclass[10pt,twocolumn,letterpaper]{article}

\usepackage{icb}
\usepackage{times}
\usepackage{epsfig}
\usepackage{graphicx}
\usepackage{amsmath}
\usepackage{amssymb}
\usepackage{multirow}

\icbfinalcopy 

\ificbfinal\fi
\begin{document}

\title{Deep Sketch-Photo Face Recognition Assisted by Facial Attributes}

\author{Seyed Mehdi Iranmanesh, Hadi Kazemi, Sobhan Soleymani, Ali Dabouei, Nasser M. Nasrabadi\\
West Virginia University\\
{\tt\small \{seiranmanesh, hakazemi, ssoleyma, ad0046\}@mix.wvu.edu, \{nasser.nasrabadi\}@mail.wvu.edu}}
\maketitle
\thispagestyle{empty}

\begin{abstract}
In this paper, we present a deep coupled framework to address the problem of matching sketch image against a gallery of mugshots. Face sketches have the essential information about the spatial topology and geometric details of faces while missing some important facial attributes such as ethnicity, hair, eye, and skin color. We propose a coupled deep neural network architecture which utilizes facial attributes in order to improve the sketch-photo recognition performance. The proposed Attribute-Assisted Deep Convolutional Neural Network (AADCNN) method exploits the facial attributes and leverages the loss functions from the facial attributes identification and face verification tasks in order to learn rich discriminative features in a common embedding subspace. The facial attribute identification task increases the inter-personal variations by pushing apart the embedded features extracted from individuals with different facial attributes,  while  the verification  task  reduces  the  intra-personal  variations  by pulling together all the features that are related to one person. The learned discriminative features  can  be  well generalized to new identities not seen in the training data. The proposed architecture is able to make full use of the sketch and complementary facial attribute information to train a deep model compared to the conventional sketch-photo recognition methods. Extensive experiments are performed on composite (E-PRIP) and semi-forensic (IIIT-D semi-forensic) datasets. The results show the superiority of our method compared to the state-of-the-art models in sketch-photo recognition algorithms. 
\end{abstract}

\section{Introduction}
 
 Face sketch recognition is an important problem when the photo of a suspect is not available or is captured with very poor quality. A face sketch is usually drawn by a forensic artist~\cite{1} or facial software~\cite{2} based on the information provided by a victim, or an eye-witness. Therefore, the generated sketch using the provided description of the victim is the only clue to identify the victim. An automatic matching method is necessary to identify a suspect accurately via searching the law enforcement face database or surveillance cameras using only the sketch of the suspect. The sketch recognition problem has been extensively studied in recent years~\cite{37}. Existing approaches can be divided in four different categories; hand-drawn viewed sketches, hand-drawn semi-forensic sketch, hand-drawn forensic sketch and software-generated composite sketch~\cite{3}. Due to the large phenomenological gap between sketch and photo domains, sketch recognition problem still remains a challenging task.    
 
 Forensic or composite sketches contain limited information such as a rough spatial topology of the suspect face and lack of some complementary information such as skin color, ethnicity, or hair color are noticeable. In addition, sketch recognition problems mainly focus on single sketch which can be unreliable in real-world situations. This unreliability can lead to a false identification~\cite{4}. In forensics investigation multiple sources of information such as verbal description of multiple witnesses or the verbal description and poor video surveillance can be utilized to enhance the performance of suspect identification~\cite{5,6}.
 
 In general there are two classical ways to solve the sketch recognition problem. First approach namely generative methods transfer one of the modalities (either sketch or photo) to the other before matching~\cite{8,9}. In the second approach, the discriminative methods utilize feature descriptors such as the scale-invariant feature transform (SIFT)~\cite{10}, Weber's local descriptor (WLD)~\cite{11}, and multi-scale local binary pattern (MLBP)~\cite{12}. The main drawbacks of these feature descriptors is that they might not be the optimal features for the task of sketch-photo recognition. To compensate for this, some other methods in the literature propose to extract modality-invariant features~\cite{36,14}. 
 
 Recently, deep learning methods have been widely utilized in face recognition and other classification problems~\cite{15,38,40,41,42, 43, 45} instead of classical methods~\cite{44,39}. These methods, can also be employed for the task of sketch-photo recognition problem by learning the relationship between the two modalities. However, the problem of sketch recognition is more challenging compared to the classical face recognition problem from the deep learning point of view. The reason behind this lies not only in the heterogeneous nature of sketch and photo modalities but also the lack of large databases in order to avoid over-fitting and local minima. For example, most of the datasets contain only one sketch per subject which makes it very challenging for a deep model to learn the robust features~\cite{16}. To avoid this, many deep techniques utilize relatively shallow model or train the network only on the photo modality~\cite{17}.
 
 Recently, in the literature soft biometric traits have been utilized jointly with hard biometrics (face photo) for different tasks such as person identification or face recognition~\cite{18}. In fact, using facial attributes in conjunction with sketch would be more advantageous since some attributes such as eye color, hair color, skin color, and ethnicity do not exist in sketch and could be considered as the complementary information. Moreover, some attributes such as wearing a hat or eyeglasses can be utilized as an auxiliary information to narrow down the suspect in the databases more accurately. Recent approaches on sketch recognition problem have mainly focused on closing the gap between the two domains of sketch and photo and use of soft biometrics has not been investigated adequately. In~\cite{19} an approach was proposed to directly use facial attributes in suspect identification without using the sketch. ~\cite{20} used race and gender to narrow down the galley of mugshots for faster and more accurate matching. Mittal et. al.~\cite{3} fused multiple sketches of a suspect to increase the accuracy of their algorithm. They also employed some soft biometric traits such as gender, ethnicity, and skin color to reorder the ranked list of the suspects. Ouyang et al.~\cite{7} introduced a framework to combine the facial attributes with low-level features to fill the gap between sketch and photo modalities.       
 
 In this paper, we propose an attribute-assisted sketch recognition framework which uses relevant facial attributes, provided by a victim, to enhance the performance of our deep sketch recognition method. Our approach simultaneously learns a common embedding features of sketch and photo image by minimizing two supervisory loss functions, namely the facial attributes identification and sketch-photo verification loss functions (tasks). Attribute identification task classifies photo and attribute assisted sketch images into a set of facial attributes, while verification task is to classify a pair of sketch-photo as belonging to the same person or not. The attribute identification loss is trying to pull the common features of photo and attribute assisted sketch closer in the shared latent subspace if they belong to the same set of attributes and push them apart if they belong to two different sets of attributes. Therefore, the learned features contain rich variations and can classify the photos and sketches to the classes containing the same sets of attributes in a latent feature subspace. 
 
 In summary, the main contributions of this paper include the following: 
 
 1- We propose a novel deep learning approach utilizing the facial attributes to improve sketch-photo recognition performance.

 2- We introduce a joint loss function which is based on an identification-verification framework in which the identification part is responsible for the facial attribute classification and the verification part is responsible for creating a common embedding subspace between the sketch and photo modalities. This loss function helps the proposed coupled deep architecture to produce a more discriminative embedding subspace which leads to a better sketch-photo recognition performance.

 3- The proposed method is able to fuse textural information of forensic sketches and complementary facial attributes such as skin color and hair color implicitly. 
 
 The remainder of this paper is organized as follows: Section 2 introduces the motivation and presents the methodology for the proposed sketch-photo recognition 
 framework. Section 3 gives comprehensive experimental results and analysis. Finally, in Section 4, conclusions
 are drawn.

\begin{figure*}
	\begin{center}
		\includegraphics[width=1\linewidth]{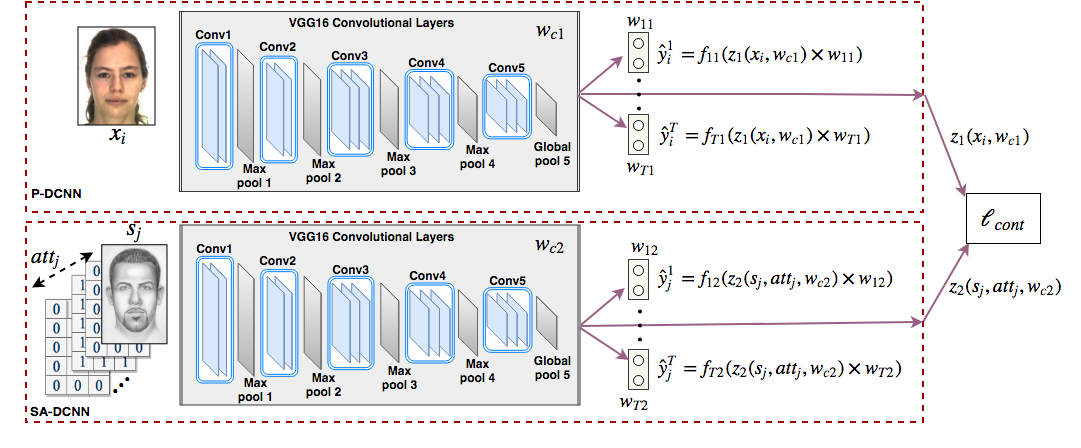}
		
	\end{center}
	\caption{Attribute-assisted deep convolutional neural network. P-DCNN (upper network) and SA-DCNN (lower network) embed the photos and a pair of (sketch, attribute) into a common latent subspace. }
	\label{fig:figure3}
\end{figure*}

 \section{Methodology}

The network parameters in the proposed framework are learned by minimizing two supervisory loss functions namely the losses due to the sketch-photo verification and facial attribute identification tasks. In the following we describe these two supervisory loss functions in details:

\subsection{Sketch-Photo Verification Task:}

\begin{figure*}
	\begin{center}
		\includegraphics[width=1\linewidth]{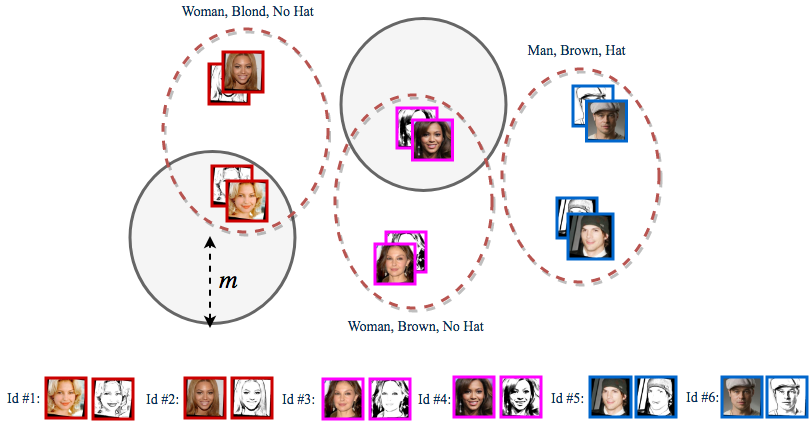}
		
	\end{center}
	\caption{Visualization of the common latent subspace by leveraging facial attributes classification and verification loss functions simultaneously. Solid circles represent the contrastive margin in the embedding domain and the dashed circles depict the attributes classification. For the sake of clarity the contrastive margin is depicted for two $Ids$ out of six $Ids$.}
	\label{fig:figure4}
\end{figure*}

 Sketch-photo verification is the final objective of the proposed model which is identification of the suspect sketch in a gallery of mugshots. For this reason, we coupled two VGG-16 like networks one dedicated to the photo image domain (P-DCNN) and the other one to the sketch and complementary facial attributes modalities (SA-DCNN). Each DCNN performs a non-linear transformation on the input. The ultimate goal of our proposed attribute-assisted deep convolutional neural network, as shown in Figure~\ref{fig:figure3}, is to find the global deep features representing the relationship between sketches and their corresponding images. In order to find the common embedding space between these two different modalities we coupled two VGG-16 structured networks (P-DCNN and SA-DCNN) via a contrastive loss function~\cite{24}. This function ($\ell_{cont}$) pulls the genuine pairs (i.e., a face photo image with its own corresponding sketch image) towards each other into a common latent feature subspace and push the impostor pairs (i.e., a photo image with sketch image from another subject) apart from each other (see Fig.~\ref{fig:figure4}). Similar to~\cite{24}, the contrastive loss is of the form:
\begin{align}
\ell_{cont}&(z_1(x_i),z_2(s_j,att_j),y_{cont})=\\ \nonumber &(1-y_{cont})L_{gen}(D(z_1(x_i),z_2(s_j,att_j))+\\ \nonumber & y_{cont}L_{imp}(D(z_1(x_i),z_2(s_j,att_j)) \; ,
\label{label-5}
\end{align}

\noindent where $x_i$ is the input for the P-DCNN (i.e., a photo image), and $(s_j, att_j)$ is the input for the SA-DCNN (i.e., an sketch image with its corresponding attributes provided by the eye witness). $y_{cont}$ is a binary label, $L_{gen}$ and $L_{imp}$ represent the partial loss functions for the genuine and impostor pairs, respectively. $z_1 $ and $z_2 $ are the DNN-based embedding functions, which transform $x_i$ and $(s_j,att_j)$ into a common latent embedding subspace, respectively, and $D(z_1(x_i),z_2(s_j,att_j))$ indicates the Euclidean distance between the embedded data in the common feature subspace. The binary label, $y_{cont}$, is assigned a value of 0 when both modalities, i.e., photo and sketch, form a genuine pair, or, equivalently, the inputs are from the same subject. On the contrary, when the inputs are from different identities, which means they form an impostor pair, $y_{cont}$ is equal to 1. In addition, $L_{gen}$ and $L_{imp}$ are defined as follows: 
\begin{equation}
\begin{split}
L_{gen}(D(z_1(x_i),z_2(s_j,att_j)))= &\dfrac{1}{2}  D(z_1(x_i),z_2(s_j,att_j))^2 \\
& \text{for} \quad\quad y_i=y_j\;,
\label{label-6}
\end{split}
\end{equation}
\begin{align}
& L_{imp}(D(z_1(x_i), z_2(s_j, att_j)))=\\ \nonumber
& \dfrac{1}{2} \:  max(0,m-D(z_1(x_i),z_2(s_j, att_j)))^2\quad   \text{for} \quad y_i \neq y_j\; .
\label{label-7} 
\end{align}

Therefore, the total loss function for the training dataset can be written as:
\begin{equation}
\begin{split}
L_1 &= 1/N^2 \displaystyle\sum_{i=1}^{N}\displaystyle\sum_{j=1}^{N} \ell_{cont}(z_1(x_i), z_2(s_j,att_j), y_{cont})  \;,
\label{label-8}
\end{split}
\end{equation}

\noindent where $N$ is the number of samples. It should be noted that the contrastive loss function~\cite{24} considers the subjects' labels inherently. Therefore, it has the ability to find a discriminative embedding space by employing the data labels in contrast to some other metrics such as the Euclidean distance. This discriminative embedding space would be useful in identifying an sketch probe against a gallery of mugshots. However, in our framework we incorporate the facial attribute identification task in addition to the contrastive function to make the embedding space more discriminative. The facial attributes identification task assigns each sketch or image domain to a set of attributes. The attributes are predicted using both the P-DCNN and SA-DCNN networks in a multi-tasking manner. In the following subsection, we describe the multi-tasking problem in the context of attribute prediction.

\subsection{Multi-Attribute Prediction and Identification Task:}

The objective of this model is to predict a set of attributes using a face photo or an sketch. Therefore, in this architecture a face photo (face sketch) is presented to the network as an input and a set of attributes are predicted. Suppose the input is an image $x_i \in X$, and its class label is $y_i \in Y$ for $i=1,\dots,N$ where $N$ is the number of the training samples. Soft biometric traits, contain $T$ different facial attributes or binary class labels provided by the eye witness. Therefore, in this framework we denote them as $ y^{t}$ for $t=1,\dots,T$ . The loss function is defined as:   

\begin{equation}
\begin{split}
L_2 =  &1/N \displaystyle\sum_{i=1}^{N} \displaystyle\sum_{t=1}^{T}\ell(f_{t1}(z_1(x_i)), y_i^{t}) \; ,
\end{split}
\label{label-1}
\end{equation}

\noindent where $\ell$ is a proper loss function (e.g., cross entropy) and $f_{t1}(z_1(x_i))$ is a binary classifier for the attribute $t$ operated on the output of P-DCNN.  Learning multiple CNNs separately is not optimal since different tasks may have some hidden relationships with each other and may share some common features. This is supported by~\cite{23} where they train a CNN features for the face recognition task and they used it directly for the face attribute estimation. Therefore, our network shares a big portion of its parameters among different tasks in order to enhance the performance of the recognition task. Thus, the loss function~(\ref{label-1}) can be reformulated as follows: 

\begin{equation}
\begin{split}
L_2 &= 1/N \displaystyle\sum_{i=1}^{N} \displaystyle\sum_{t=1}^{T}\ell(f_{t1}(z_1(x_i, w_{c1})\times w_{t1}), y_i^t)\; ,
\end{split}
\label{label-2}
\end{equation}

\noindent where $\ell$ is the cross entropy loss function. $w_{c1}$ is the shared network parameters between all the tasks and $w_{t1}$ represents the remaining parameters which are assigned separately for each facial attribute task.

The same procedure is performed in the other network (SA-DCNN) with an sketch as input. However, there are some attributes such as hair color and skin color which do not exist in the sketch modality while they inherently exist in the RGB images. These are the soft biometric traits which is provided by the eye witness description. Therefore, these complementary soft biometric traits are given to the SA-DCNN network which is dedicated to sketch modality. The SA-DCNN network is also responsible to estimate a set of soft biometric attributes. It should be noted that although some of the attributes in the output are given to the network from the beginning, but this attributes are fused with sketch information through the network layers. Therefore, it is worth to estimate them accurately. Also, the set of attributes which are given to the network are not necessarily the same as the set of attributes predicted by the network.

Suppose the input is an sketch $s_j \in S$, and its class label $y_j \in Y$ for $j=1,\dots,N$ where $N$ is the number of the training samples. The facial attributes provided by the eye witness are also given to the network as an input, denoted as $att$ for the sake of clarity. The loss function will be defined as:   

\begin{equation}
\begin{split}
L_3 =  &1/N \displaystyle\sum_{j=1}^{N} \displaystyle\sum_{t=1}^{T}\ell (f_{t2}(z_2(s_j,att_j)), y_j^{t}) \; ,
\end{split}
\label{label-3}
\end{equation}

\noindent where $\ell$ is the cross entropy loss function and $f_{t2}(z_2(s_j,att_j))$ is a binary classifier for the attribute $t$ operated on the output of SA-DCNN. Here, as in P-DCNN network , we share a big portion of the network parameters among different tasks in order to enhance the performance of the recognition task. Therefore, the loss function (\ref{label-3}) can be reformulated as follows: 

\begin{equation}
\begin{split}
L_3 =  &1/N \displaystyle\sum_{j=1}^{N} \displaystyle\sum_{t=1}^{T}\ell (f_{t2}(z_2(s_j,att_j, w_{c2})\times w_{t2}), y_j^t)\; ,
\end{split}
\label{label-4}
\end{equation}

\noindent where $w_{c2}$ is the shared features between all the tasks. $w_{t2}$ represents the remaining features which are assigned separately for each soft biometric prediction task.

\subsection{Total Loss Function:}

The total loss function $L_T$ for the whole framework can be written as (See Fig.~\ref{fig:figure3}) : 

\begin{equation}
\begin{split}
L_T & = L_1 + \lambda_1 L_2 +\lambda_2 L_3 = \\&  1/N^2\displaystyle\sum_{i=1}^{N}\displaystyle\sum_{j=1}^{N} \ell_{cont}(z_1(x_i), z_2(s_j,att_j), y_{cont})  \\& + \lambda_1/N \displaystyle\sum_{i=1}^{N} \displaystyle\sum_{t=1}^{T}\ell (f_{t1}(z_1(x_i)), y_i^{t})\\ & +\lambda_2/N \displaystyle\sum_{j=1}^{N} \displaystyle\sum_{t=1}^{T}\ell (f_{t2}(z_2(s_j,att_j)), y_j^{t})\; ,
\label{label-9}
\end{split}
\end{equation}

\noindent where the first term is the sketh-photo verification and the second and the third terms are the facial attributes classification loss for the P-DCNN network and SA-DCNN network, respectively. $ \lambda_1$ and $ \lambda_2$ are the hyper-parameters which weight facial identification cost functions of the P-DCNN network and SA-DCNN network, respectively.   
As it was mentioned earlier, the contrastive loss function has the ability to find a discriminative embedding space by employing the data labels. However, due to loss functions from the facial attributes classification term for photo (\ref{label-1}) and for sketch (\ref{label-3}), minimizing $L_T$ will boost the discrimination in the common embedding domain. In another words, using just the contrastive loss it does not consider whether two subjects share similar facial attributes or not. Using the facial attribute classification, it enables the embedding space to be more discriminative from the attributes point of view. 

Consider a subject sketch with $Id \#1$ (see Fig.~\ref{fig:figure4}). The contrastive loss function causes the corresponding photos from $Id\#1$ to move closer to $Id\#1's$ sketch and other $Ids'$ photos to move farther away. Now, using the contrastive loss function in conjunction with the attribute classification makes $Id \#2$ to move closer to $Id \#1$ since they share the same set of attributes (see Fig.~\ref{fig:figure4}). In other words, it differentiates between different impostors of $Id \#1$. The same procedure is performed for the other identities during the training process. Figure~\ref{fig:figure4} visualizes the overall concept of our joint loss function. As it is depicted, jointly training the model based on verification and facial identification will lead to a more discriminative embedding subspace which considers both the facial attributes and the geometrical relationship between the forensic sketches and photos. 

During the testing phase, given a test probe with its facial attributes $(s_t,att_t)$, the proposed AADCNN method transforms it to the common latent embedding domain, $ z_2(s_t,att_t)$. In fact, after training our deep coupled network model, it has the ability to transform the photo and sketch images into a common discriminative embedding space. Therefore, the galley of the photo images is transformed to the mentioned embedding space. Eventually, the sketch image is identified, by calculating the minimum Euclidean distance between the transformed sketch prob and gallery of mugshots as follows:

\begin{equation}
\begin{split}
x_i^* = & \underset{x_i}{\operatorname{argmin}}\quad D(z_1(x_i),z_2(s_t,att_t)) 
\\& \text{for} \quad\quad i=1,2,..,M \;,
\label{label-10}
\end{split}
\end{equation}

\noindent where $ (s_t,att_t)$ is an sketch probe with its facial attribute provided by the eye witness and $x_i^*$ is the selected matching person within the gallery of mugshots of size $M$.

\section{Experiments and Evaluations}

\subsection{Implementation Details and Data Description}

In this paper, we used a VGG-16 like network~\cite{22} in our sketch-photo recognition framework. The VGG-16 neural network comprised of five major convolutional components which are connected in series. The first two components, $Conv1-64$ and $Conv2-128$ consists of the following layers: a convolutional layer, a rectified linear unit layer, a second convolutional layer, a second rectified linear unit layer, and a max pooling layer. The remaining three components contain one additional convolutional layer and a rectified linear unit layer. The only difference between our CNN network and the VGG-16 is in the last component, where the last three convolutional layers of VGG16 with the size of 512, are replaced with two convolutional layers of 256, and one convolutional layer of 64 respectively for the sake of parameter reduction. Also the network uses global pooling instead of the max pooling in the last component which results in a feature vector of size 64. The network which is dedicated to photo domain (P-DCNN) takes an RGB photo as an input and the other sketch-attribute network (SA-DCNN) gets an input consisting of multiple channels as shown in Fig.~\ref{fig:figure3}. The first channel is dedicated to a gray-scale sketch and the other channels are filled with 0 or 1 depending on the presence or absence of the attribute in the subject.

Experiment is performed using four main datasets, namely CUHK Face Sketch dataset (CUFS)~\cite{25} (containing 311 pairs), IIIT-D sketch dataset~\cite{26} (containing 238 viewed pairs, 140 semi-forensic pairs, and 190 forensic pairs), PRIP Viewed Software Generated Composite database (PRIP-VSGC)~\cite{2} (containing composite sketch and digital image pairs), extended-PRIP dataset (e-PRIP)~\cite{3}, and unviewed Memory Gap Database (MGDB)~\cite{8} (containing 100 pairs). Since we are using the facial attribute classification in our proposed method, we utilized the CelebFaces Attributes dataset (CelebA)~\cite{29} (consisting of 200 k face images along with their attribute vectors of 40 attributes such as gender, face characteristics, skin color, hair color, etc.) to initialize the network. Since the CelebA dataset does not contain the sketch images we generated a synthetic sketch by employing xDOG~\cite{30} filter on each image. Twelve facial attributes namely bald, black hair, blond hair, brown hair, gray hair, male, Asian, Indian, White, Black, eye glasses and pale skin out of 40 attributes were selected. Since none of the sketch datasets used in this paper have any facial attribute annotation, we utilized MOON~\cite{33}, which is a well-known method in facial attributes recognition, in order to annotate them.       

\begin{figure}
	\begin{center}
		\includegraphics[width=1\linewidth]{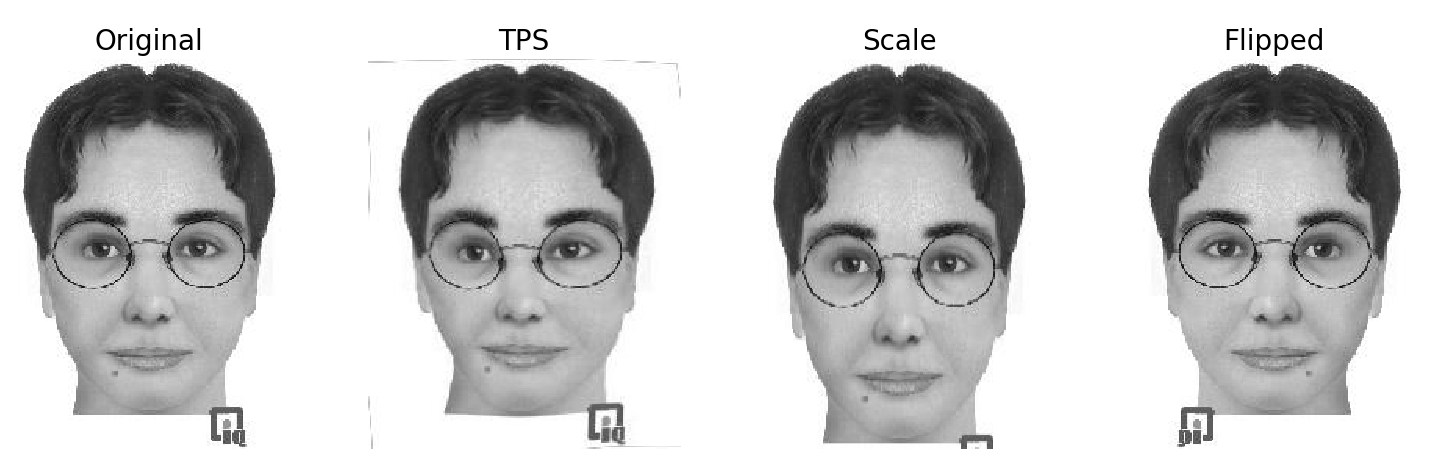}
		
	\end{center}
	\caption{A sample of different augmentation techniques.}
	\label{fig:figure_deform}
\end{figure}

We pre-trained our deep coupled architecture using synthetic sketch-photo pair from the CelebA dataset. We used the final weights to initialize the network in all of our training scenarios.
Since our coupled DNN has a large number of parameters and the size of the sketch datasets is relatively small it is prone to overfitting. In order to avoid the overfitting problem, we utilized multiple augmentation techniques namely deformation, scale and crop, and flipping. In the following we explain each method in details (see Fig.~\ref{fig:figure_deform} ). 

1- Deformation: Deforms the sketch and photos to compensate for the problem of geometrically mismatching between the sketch image and its corresponding photo. The deformation is performed by translating 25 preselected points with random magnitude and direction. 

2- Scale and crop: One of the main mismatch problems between sketch images and their corresponding images is the ratio deformation. To address this problem, this method upscale the sketches and photos to several random sizes, and then cut a $250\times200$ crop from the center of the scaled image. 

3- Flipping: In this method, the images are randomly flipped horizontally. 

During the training phase, instead of picking the impostor pairs randomely, we considered an strategy to select them. For each genuine pair, we considered four impostor pairs. Two of the impostors were selected among the subjects which are sharing the same set of facial attributes and the other two were selected among the subjects which have different sets of attributes. This selection technique made the framework to see more variant impostors and also helped to avoid the overfitting problem.   

\begin{table*}[]
	\centering
	\caption{Experimental Setup}
	\label{table:Ps}
	\begin{tabular}{|c|c|c|c|c|c|}
		\hline
		Setup Name & Testing Dataset    & Training Dataset                  	& Train Size & Gallery Size & Prob Size \\ \hline
		S1       & e-PRIP               & e-PRIP                            	& 48         & 75           & 75        \\ \hline
		S2       & e-PRIP               & e-PRIP                            	& 48         & 1500         & 75        \\ \hline
		\multirow{2}{*}{S3}  & IIIT-D Semi-forensic & \multirow{2}{*}{CUFS, IIIT-D Viewed, CUFSF, e-PRIP}    & \multirow{2}{*}{1968}       & \multirow{2}{*}{1500}         & 135       \\ \cline{2-2}\cline{6-6}
		&   MGDB Unviewed & &        &          & 100       \\ \hline
	\end{tabular}
\end{table*}
\subsection{Performance Evaluation:}

Our proposed framework identify a person of interest in the galley of mugshots utilizing a sketch probe and a set of facial attributes provided. In this section, we compare our approach with several state-of-the-art methods which are using both sketch and attributes and some other methods which are just using the sketch without using any attributes. 

In order to evaluate performance of our method and compare with other methods, three different experiments are performed. For the sake of fair comparison, the first two experiment setups are adopted from~\cite{3}. In the first experimental setup which is the baseline (S1), the database is partitioned into two parts: training which is performed on $40\%$ of the data and the remaining portion of the data is used for testing. e-PRIP dataset containing 123 subjects is used in this setup. Therefore, 48 identities are used in the training set and 75 subjects are  considered for the testing phase. Only two out of the four different composite sketch datasets utilized in~\cite{3} are public at the time of writing this paper. These two public datasets were created by Identi-Kit tool, and FACES tool and were used by Asian and Indian artists respectively. In the second experimental setup, called S2, the gallery is extended to 1500 subjects. In this paper, the gallery is expanded utilizing WVU Multi-Modal~\cite{32}, IIIT-D sketch, Multiple Encounter Dataset (MEDS)~\cite{13}, and CUFS datasets. The facial attributes of the extended galley are obtained using MOON~\cite{33}. The training and probe datasets are the same as S1. The purpose of this experiment is to assess the robustness of the proposed method with a relatively large number of subject candidates. Finally, the method is evaluated on an unseen dataset. In this experimental setup (S3), we trained the network on IIIT-D Viewed, CUFS, and e-PRIP datasets and then tested it on IIIT-D Semi-forensic pairs and MGDB Unviewed.  This setup represents the level of dependency of the network on the sketch styles in the training datasets. Table~\ref{table:Ps} shows different scenarios and the size of training set, prob and gallery for each scenario.   

In the experiments, different values were selected for $\lambda_1$ and $\lambda_2$. We report our best results which belong to $\lambda_1$=$\lambda_2$=1.  The evaluation performance is validated using ten fold random cross validation and the results are compared with the state-of-the-art approaches. 

\subsection{Results:}

\begin{figure}
	\begin{center}
		\includegraphics[width=1\linewidth]{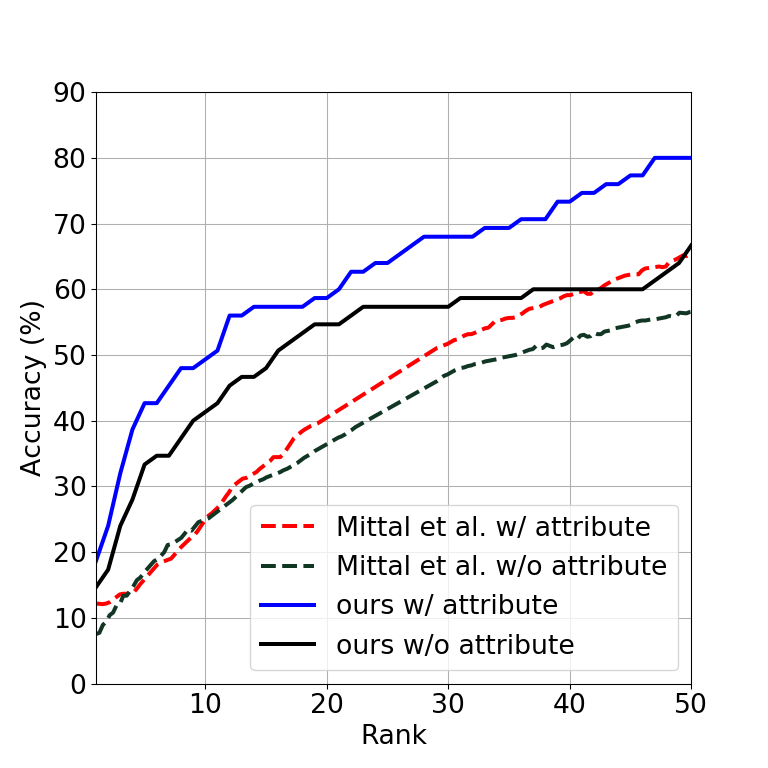}
		
	\end{center}
	\caption{CMC curves of our proposed framework versus Mittal et al. algorithms~\cite{3} in the extended gallery experimental setup (S2) for the Indian dataset  }
	\label{fig:figure6}
\end{figure}

In~\cite{3}, they propose an approach called attribute feedback to study the effect of facial attributes on their recognition system. They reported the rank 10 accuracy of $58.4\% $ and $ 53.1\%$ for the prob sketches generated by the Indian (Faces) and Asian (Identi-Kit) artists, respectively. Another approach called SGR-DA~\cite{34} utilizes the sketch modality without using the facial attribute information. They reported the rank 10 accuracy of $70\% $ on the Identi-Kit dataset. On the other hand, our proposed approach accuracy was $76.4\%$ and $72.3\%$ on the Faces and Identi-Kit, respectively. We also consider a baseline version of our proposed method which is only based on the contrastive loss function and does not consider the facial attributes. This way, we could observe the benefit of utilizing the facial attributes in our framework. The baseline network has an accuracy of $69.1\%$ and $67.6\%$, on Faces and Identi-Kit datasets, respectively. The results demonstrate that our method outperforms all the previous methods in the literature and also express the effectiveness of our framework in utilizing the facial attributes compare to the baseline. It should be noted that, the baseline framework outperform the state-of-the-art methods except SGR-DA~\cite{34} which support the superiority of deep models over the shallow models (see Table~\ref{table:P1}).        

\begin{figure}
	\begin{center}
		\includegraphics[width=1\linewidth]{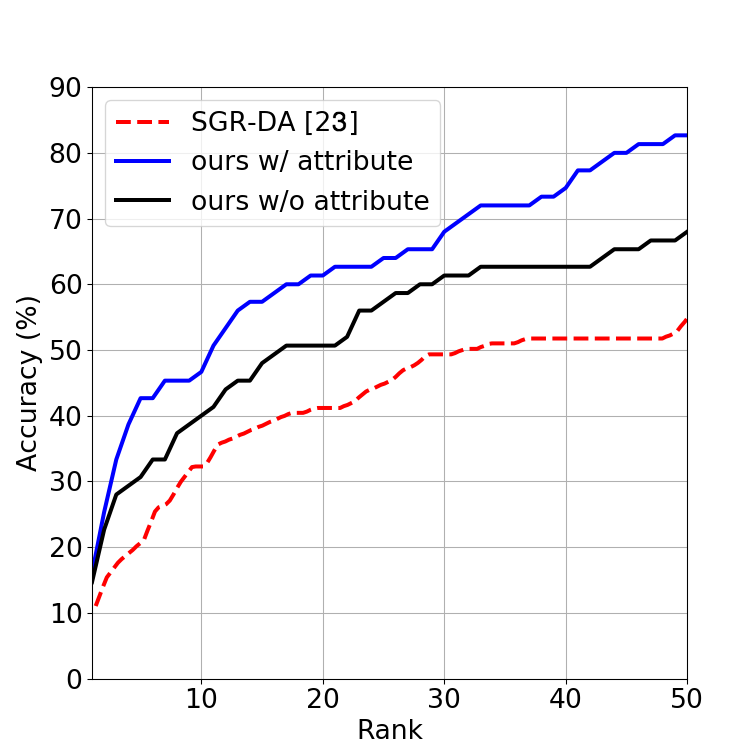}
		
	\end{center}
	\caption{CMC curves of our proposed framework versus SGR-DA algorithm~\cite{34} in the extended gallery experimental setup (S2) for the Identi-Kit dataset}
	\label{fig:figure7}
\end{figure}

To evaluate the effectiveness of our proposed method by using a relatively large galley of mugshots, the same experiments were performed on the extended experimental setup (S2). Figure~\ref{fig:figure6} shows the results of our method as well as the other methods for the extended galley of 1500 subjects. The results depicts that our approach outperforms the method in~\cite{3} by nearly $14\%$ for rank 50 which shows the robustness of our algorithm utilizing the facial attributes. We compared our method with SGR-DA~\cite{34} for the Identi-Kit dataset, since~\cite{3} does not provide the results on this dataset. Figure~\ref{fig:figure7} shows the superiority of our proposed method compared to SGR-DA. As shown in Fig.~\ref{fig:figure7}, although SGR-DA outperformed our baseline network in S1 scenario (see Table~\ref{table:P1}), its results were not as promising as our proposed method in the extended experimental setup (S2). Also, our attribute-assisted method outperformed our baseline method to support the effectiveness of utilizing the attributes in relatively large gallery of mugshots as well. 

\begin{table}[]
	\centering
	\caption{Rank-10 identification accuracy (\%) on the e-PRIP composite sketch database (S1 experimental setup).}
	\label{table:P1}
	\begin{tabular}{|l|c|c|}
		\hline
		\textbf{Algorithm}       & \textbf{Faces (In)} & \textbf{IdentiKit (As)} \\ \hline
		Mittal et al. \cite{35} & 53.3 $\pm$ 1.4      & 45.3 $\pm$ 1.5              \\ \hline
		Mittal et al. \cite{17} & 60.2 $\pm$ 2.9      & 52.0 $\pm$ 2.4              \\ \hline
		Mittal et al. \cite{3} & 58.4 $\pm$ 1.1      & 53.1 $\pm$ 1.0              \\ \hline
		SGR-DA \cite{34} & -     & 70              \\ \hline
		Ours without attributes  & 69.1 $\pm$ 1.5           & 67.6 $\pm$ 1.9                   \\ \hline
		Ours with attributes     & \textbf{76.4 $\pm$ 1.2}   &   \textbf{72.3 $\pm$ 0.8}       \\ \hline
	\end{tabular}
\end{table}

\begin{figure}
	\begin{center}
		\includegraphics[width=1\linewidth]{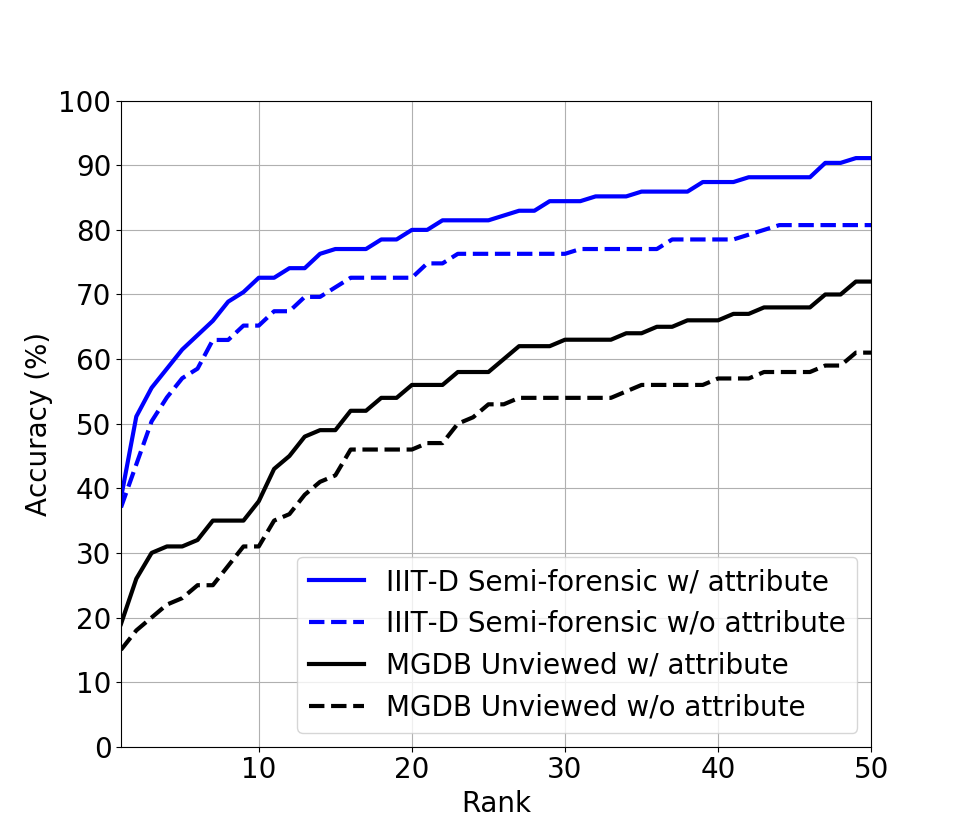}
		
	\end{center}
	\caption{CMC curves of our proposed framework versus our baseline framework (without using attributes) for experimental setup (S3). The results support the robustness of our approach to different sketch styles.}
	\label{fig:figure8}
\end{figure}

Eventually, we evaluated the robustness of our proposed method in S3 experimental setup in which the network is trained on more than 1900 sketch-photo pairs and is tested on two unseen datasets, namely MGDB Unviewed and IIIT-D Semi-forensic datasets. In this scenario the gallery of mugshots was also extended to 1500. We repeated this experimental scenario for our baseline method which is not utilizing the facial attributes. As shown in Fig.~\ref{fig:figure8}, the proposed method showed a better performance in this scenario on both datasets compared to the baseline method indicating the advantage of facial attributes in the proposed method on unseen datasets. 

\section{Conclusion}
We have introduced a novel approach to exploit facial attributes information for the purpose of sketch-photo recognition. The proposed network is capable of transforming the photo and sketch modalities into a common discriminative embedding subspace. We have proposed to use coupled deep neural network with facial attributes provided by eye witnesses. We simultaneously minimize the cost functions due to the facial attribute identification as well as the sketch-photo verification in order to increase inter-personal variations between different subjects with different sets of facial attributes and reducing intra-personal variations in the latent feature subspace. The combination of the two cost functions leads to a significantly more discriminative embedding subspace compared to the subspace that is created by either one of them. We compared our method with state-of-the-art sketch-photo recognition methods and showed the superiority of our method over them.

{\small
\bibliographystyle{ieee}
\bibliography{ICB}

\begin{thebibliography}{10}\itemsep=-1pt

\bibitem{32}
Biometrics and identification innovation center, wvu multi- modal dataset.
  available at http://biic.wvu.edu/,.

\bibitem{39}
D.~Alhelal, K.~A. Aboalayon, M.~Daneshzand, and M.~Faezipour.
\newblock Fpga-based denoising and beat detection of the ecg signal.
\newblock In {\em Systems, Applications and Technology Conference (LISAT), 2015
  IEEE Long Island}, pages 1--5. IEEE, 2015.

\bibitem{5}
L.~Best-Rowden, H.~Han, C.~Otto, B.~F. Klare, and A.~K. Jain.
\newblock Unconstrained face recognition: Identifying a person of interest from
  a media collection.
\newblock {\em IEEE Transactions on Information Forensics and Security},
  9(12):2144--2157, 2014.

\bibitem{26}
H.~S. Bhatt, S.~Bharadwaj, R.~Singh, and M.~Vatsa.
\newblock Memetic approach for matching sketches with digital face images.
\newblock Technical report, 2012.

\bibitem{11}
H.~S. Bhatt, S.~Bharadwaj, R.~Singh, and M.~Vatsa.
\newblock Memetically optimized mcwld for matching sketches with digital face
  images.
\newblock {\em IEEE Transactions on Information Forensics and Security},
  7(5):1522--1535, 2012.

\bibitem{44}
A.~Broumand, M.~S. Esfahani, B.-J. Yoon, and E.~R. Dougherty.
\newblock Discrete optimal bayesian classification with error-conditioned
  sequential sampling.
\newblock {\em Pattern Recognition}, 48(11):3766--3782, 2015.

\bibitem{24}
S.~Chopra, R.~Hadsell, and Y.~LeCun.
\newblock Learning a similarity metric discriminatively, with application to
  face verification.
\newblock {\em IEEE Conference on Computer Vision and Pattern Recognition
  (CVPR)}, 1:539--546, 2005.

\bibitem{40}
A.~Dabouei, H.~Kazemi, S.~M. Iranmanesh, J.~Dawson, and N.~M. Nasrabadi.
\newblock Fingerprint distortion rectification using deep convolutional neural
  networks.
\newblock {\em arXiv preprint arXiv:1801.01198}, 2018.

\bibitem{18}
A.~Dantcheva, P.~Elia, and A.~Ross.
\newblock What else does your biometric data reveal? a survey on soft
  biometrics.
\newblock volume~11, pages 441--467. IEEE, 2016.

\bibitem{13}
A.~P. Founds, N.~Orlans, W.~Genevieve, and C.~I. Watson.
\newblock Nist special databse 32-multiple encounter dataset ii (meds-ii).
\newblock Technical report, 2011.

\bibitem{12}
C.~Galea and R.~A. Farrugia.
\newblock Face photo-sketch recognition using local and global texture
  descriptors.
\newblock In {\em Signal Processing Conference (EUSIPCO), 2016 24th European},
  pages 2240--2244. IEEE, 2016.

\bibitem{16}
C.~Galea and R.~A. Farrugia.
\newblock Forensic face photo-sketch recognition using a deep learning-based
  architecture.
\newblock {\em IEEE Signal Processing Letters}, 24(11):1586--1590, 2017.

\bibitem{6}
L.~Gibson.
\newblock {\em Forensic art essentials: a manual for law enforcement artists}.
\newblock Academic Press, 2010.

\bibitem{2}
H.~Han, B.~F. Klare, K.~Bonnen, and A.~K. Jain.
\newblock Matching composite sketches to face photos: A component-based
  approach.
\newblock {\em IEEE Transactions on Information Forensics and Security},
  8(1):191--204, 2013.

\bibitem{14}
D.-A. Huang and Y.-C.~F. Wang.
\newblock Coupled dictionary and feature space learning with applications to
  cross-domain image synthesis and recognition.
\newblock In {\em Computer Vision (ICCV), 2013 IEEE International Conference
  on}, pages 2496--2503. IEEE, 2013.

\bibitem{43}
S.~M. Iranmanesh, A.~Dabouei, H.~Kazemi, and N.~M. Nasrabadi.
\newblock Deep cross polarimetric thermal-to-visible face recognition.
\newblock {\em arXiv preprint arXiv:1801.01486}, 2018.

\bibitem{37}
H.~Kazemi, S.~Soleymani, A.~Dabouei, M.~Iranmanesh, and N.~M. Nasrabadi.
\newblock Attribute-centered loss for soft-biometrics guided face sketch-photo
  recognition.
\newblock {\em arXiv preprint arXiv:1804.03082}, 2018.

\bibitem{10}
B.~Klare and A.~K. Jain.
\newblock Sketch-to-photo matching: a feature-based approach.
\newblock In {\em Biometric Technology for Human Identification VII}, volume
  7667, page 766702. International Society for Optics and Photonics, 2010.

\bibitem{20}
B.~Klare, Z.~Li, and A.~K. Jain.
\newblock Matching forensic sketches to mug shot photos.
\newblock {\em IEEE Transactions on Pattern Analysis and Machine Intelligence},
  33(3):639--646, 2011.

\bibitem{36}
B.~F. Klare and A.~K. Jain.
\newblock Heterogeneous face recognition using kernel prototype similarities.
\newblock {\em IEEE transactions on pattern analysis and machine intelligence},
  35(6):1410--1422, 2013.

\bibitem{19}
B.~F. Klare, S.~Klum, J.~C. Klontz, E.~Taborsky, T.~Akgul, and A.~K. Jain.
\newblock Suspect identification based on descriptive facial attributes.
\newblock In {\em Biometrics (IJCB), 2014 IEEE International Joint Conference
  on}, pages 1--8. IEEE, 2014.

\bibitem{29}
Z.~Liu, P.~Luo, X.~Wang, and X.~Tang.
\newblock Deep learning face attributes in the wild.
\newblock In {\em Proceedings of the IEEE International Conference on Computer
  Vision}, pages 3730--3738, 2015.

\bibitem{35}
P.~Mittal, A.~Jain, G.~Goswami, R.~Singh, and M.~Vatsa.
\newblock Recognizing composite sketches with digital face images via ssd
  dictionary.
\newblock In {\em Biometrics (IJCB), 2014 IEEE International Joint Conference
  on}, pages 1--6. IEEE, 2014.

\bibitem{3}
P.~Mittal, A.~Jain, G.~Goswami, M.~Vatsa, and R.~Singh.
\newblock Composite sketch recognition using saliency and attribute feedback.
\newblock {\em Information Fusion}, 33:86--99, 2017.

\bibitem{17}
P.~Mittal, M.~Vatsa, and R.~Singh.
\newblock Composite sketch recognition via deep network-a transfer learning
  approach.
\newblock In {\em Biometrics (ICB), 2015 International Conference on}, pages
  251--256. IEEE, 2015.

\bibitem{38}
S.~Motiian, Q.~Jones, S.~Iranmanesh, and G.~Doretto.
\newblock Few-shot adversarial domain adaptation.
\newblock In {\em Advances in Neural Information Processing Systems}, pages
  6670--6680, 2017.

\bibitem{7}
S.~Ouyang, T.~Hospedales, Y.-Z. Song, and X.~Li.
\newblock Cross-modal face matching: beyond viewed sketches.
\newblock In {\em Asian Conference on Computer Vision}, pages 210--225.
  Springer, 2014.

\bibitem{8}
S.~Ouyang, T.~M. Hospedales, Y.-Z. Song, and X.~Li.
\newblock Forgetmenot: Memory-aware forensic facial sketch matching.
\newblock In {\em Proceedings of the IEEE Conference on Computer Vision and
  Pattern Recognition}, pages 5571--5579, 2016.

\bibitem{34}
C.~Peng, X.~Gao, N.~Wang, and J.~Li.
\newblock Sparse graphical representation based discriminant analysis for
  heterogeneous face recognition.
\newblock {\em arXiv preprint arXiv:1607.00137}, 2016.

\bibitem{4}
C.~Peng, N.~Wang, X.~Gao, and J.~Li.
\newblock Face recognition from multiple stylistic sketches: Scenarios,
  datasets, and evaluation.
\newblock In {\em European Conference on Computer Vision}, pages 3--18.
  Springer, 2016.

\bibitem{33}
E.~M. Rudd, M.~G{\"u}nther, and T.~E. Boult.
\newblock Moon: A mixed objective optimization network for the recognition of
  facial attributes.
\newblock In {\em European Conference on Computer Vision}, pages 19--35.
  Springer, 2016.

\bibitem{15}
F.~Schroff, D.~Kalenichenko, and J.~Philbin.
\newblock Facenet: A unified embedding for face recognition and clustering.
\newblock In {\em Proceedings of the IEEE conference on computer vision and
  pattern recognition}, pages 815--823, 2015.

\bibitem{22}
K.~Simonyan and A.~Zisserman.
\newblock Very deep convolutional networks for large-scale image recognition.
\newblock {\em arXiv preprint arXiv:1409.1556}, 2014.

\bibitem{41}
S.~Soleymani, A.~Dabouei, H.~Kazemi, J.~Dawson, and N.~M. Nasrabadi.
\newblock Multi-level feature abstraction from convolutional neural networks
  for multimodal biometric identification.
\newblock In {\em 24th International Conference on Pattern Recognition (ICPR)},
  2018.

\bibitem{42}
S.~Soleymani, A.~Torfi, J.~Dawson, and N.~M. Nasrabadi.
\newblock Generalized bilinear deep convolutional neural networks for
  multimodal biometric identification.
\newblock In {\em IEEE International Conference on Image Processing (ICIP)},
  2018.

\bibitem{25}
X.~Tang and X.~Wang.
\newblock Face sketch synthesis and recognition.
\newblock In {\em Computer vision, 2003. proceedings. ninth ieee international
  conference on}, pages 687--694. IEEE, 2003.

\bibitem{45}
A.~Torfi, S.~M. Iranmanesh, N.~Nasrabadi, and J.~Dawson.
\newblock 3d convolutional neural networks for cross audio-visual matching
  recognition.
\newblock {\em IEEE Access}, 5:22081--22091, 2017.

\bibitem{1}
X.~Wang and X.~Tang.
\newblock Face photo-sketch synthesis and recognition.
\newblock {\em IEEE Transactions on Pattern Analysis and Machine Intelligence},
  31(11):1955--1967, 2009.

\bibitem{9}
Y.~Wang, L.~Zhang, Z.~Liu, G.~Hua, Z.~Wen, Z.~Zhang, and D.~Samaras.
\newblock Face relighting from a single image under arbitrary unknown lighting
  conditions.
\newblock {\em IEEE Transactions on Pattern Analysis and Machine Intelligence},
  31(11):1968--1984, 2009.

\bibitem{30}
H.~Winnem{\"o}ller, J.~E. Kyprianidis, and S.~C. Olsen.
\newblock Xdog: an extended difference-of-gaussians compendium including
  advanced image stylization.
\newblock {\em Computers \& Graphics}, 36(6):740--753, 2012.

\bibitem{23}
Y.~Zhong, J.~Sullivan, and H.~Li.
\newblock Face attribute prediction using off-the-shelf cnn features.
\newblock In {\em Biometrics (ICB), 2016 International Conference on}, pages
  1--7. IEEE, 2016.

\end{thebibliography}
}

\end{document}